\documentclass[11pt,a4paper]{article}
\usepackage[hyperref]{naaclhlt2019}
\pdfoutput=1

\usepackage{latexsym}
\usepackage{amsfonts}
\usepackage{times}
\usepackage{latexsym}
\usepackage{amsmath}
\usepackage{multirow}
\usepackage{url}
\usepackage{comment}
\usepackage{amsmath}
\usepackage{multicol}
\usepackage[pdftex]{graphicx}
\usepackage{svg}
\usepackage{bbm}
\usepackage{siunitx}

\DeclareMathOperator*{\argmax}{arg\,max}

\newcommand\LinDataset{{\sc NytFb-280k}}
\newcommand\RiedelDataset{{\sc NytFb-68k}}
\newcommand\OurSystem{{\sc PcnnNmar}}

\aclfinalcopy 

\title{Structured Minimally Supervised Learning for Neural \\ Relation Extraction}

\author{Fan Bai \and Alan Ritter\\
  Department of Computer Science and Engineering \\
  The Ohio State University \\
  Columbus, OH \\
  {\tt \{bai.313, ritter.1492\}@osu.edu}}

\date{}

\begin{document}
\maketitle
\begin{abstract}
We present an approach to minimally supervised relation extraction that combines the benefits of learned representations and structured learning, and accurately predicts sentence-level relation mentions given only proposition-level supervision from a KB.  By explicitly reasoning about missing data during learning, our approach enables large-scale training of 1D convolutional neural networks while mitigating the issue of label noise inherent in distant supervision.  Our approach achieves state-of-the-art results on minimally supervised sentential relation extraction, outperforming a number of baselines, including a competitive approach that uses the attention layer of a purely neural model.\footnote{ Our code and data are publicly available on Github: \url{https://github.com/bflashcp3f/PCNN-NMAR}} 
\end{abstract}

\section{Introduction}
Recent years have seen significant progress on tasks such as object detection, automatic speech recognition and machine translation.  These performance advances are largely driven by the application of neural network methods on large, high-quality datasets. 
In contrast, traditional datasets for relation extraction are based on expensive and time-consuming human annotation \cite{ACE} and are therefore relatively small. 
Distant supervision \cite{mintz2009distant}, a technique which uses existing knowledge bases such as Freebase or Wikipedia as a source of weak supervision, enables learning from large quantities of unlabeled text and is a promising approach for scaling up.  Recent work has shown promising results from large-scale training of neural networks for relation extraction \cite{toutanova2015representing,zeng-EtAl:2015:EMNLP}.




There are, however, significant challenges due to the inherent noise in distant supervision.
For example, Riedel et al. \shortcite{riedel10modeling} showed that, when learning using distant supervision from a knowledge base, the portion of mis-labeled examples can vary from 13\% to 31\%. 
To address this issue, another line of work has explored {\em structured} learning methods that introduce latent variables.  An example is MultiR \cite{hoffmann-EtAl:2011:ACL-HLT2011}, which is based on a joint model of relations between entities in a knowledge base and those mentioned in text.  This structured learning approach has a number of advantages; for example, by integrating inference into the learning procedure it has the potential to overcome the challenge of missing facts by ignoring the knowledge base when mention-level classifiers have high confidence \cite{ritter13,xu2013filling}.
Prior work on structured learning from minimal supervision has leveraged sparse feature representations, however, and has therefore not benefited from learned representations, which have recently achieved state-of-the-art results on a broad range of NLP tasks.

In this paper, we present an approach that combines the benefits of structured and neural methods for minimally supervised relation extraction.  Our proposed model learns sentence representations that are computed by a 1D convolutional neural network \cite{collobert2011natural} and are used to define potentials over latent relation mention variables.  These mention-level variables are related to observed facts in a KB using a set of deterministic factors, followed by pairwise potentials that encourage agreement between extracted propositions and observed facts, but also enable inference to override these soft constraints during learning, allowing for the possibility of missing information.  Because marginal inference is intractable in this model, a MAP-based approach to learning is applied \cite{taskar2004max}.  

Our approach is related to recent work structured learning with end-to-end learned representations, including Structured Prediction Energy Networks (SPENs) \cite{belanger2016structured}; the key differences are the application to minimally supervised relation extraction and the inclusion of latent variables with deterministic factors, which we demonstrate enables effective learning in the presence of missing data in distant supervision.
Our proposed method achieves state-of-the-art results on minimally supervised sentential relation extraction, outperforming a number of baselines including one that leverages the attention layer of a purely neural model \cite{lin-EtAl:2016:P16-1}.


\section{A Latent Variable Model for Neural Relation Extraction}
In this section we present our model, which combines continuous representations with structured learning.  We first review the problem setting and introduce notation, next we present our approach to extracting feature representations which is based on the piecewise convolutional neural network (PCNN) model of Zeng et. al. \shortcite{zeng-EtAl:2015:EMNLP} and includes positional embeddings \cite{collobert2011natural}.
Finally we describe how this can be combined with structured latent variable models that reason about overlapping relations and missing data during learning.



\subsection{Assumptions and Problem Formulation}
\label{sec:challenges}
Given a set of sentences, $\mathbf{s} = s_1, s_2 \ldots, s_n$ that mention a pair of knowledge base entities $e_1$ and $e_2$ (dyad), our goal is to predict which relation, $r$, is mentioned between $e_1$ and $e_2$ in the context of each sentence, represented by a set of hidden variables, $\mathbf{z} = z_1, z_2, \ldots z_n$.  Relations are selected from a fixed set drawn from a knowledge base, in addition to {\em NA} (no relation).  Minimally supervised learning is more difficult than supervised relation extraction, because we do not have direct access to relation labels on the training sentences.  Instead, during learning, we are only provided with information about what relations hold between $e_1$ and $e_2$ according to the KB.  The problem is further complicated by the fact that most KBs are highly incomplete (this is the reason we want to extend them by extracting information from text in the first place), which effectively leads to false-negatives during learning.  Furthermore, there are many overlapping relations between dyads, so it is easy for a model trained using minimal supervision from a KB to confuse these relationships.  All of these issues are addressed to some degree by the structured learning approach that we present in Section \ref{sec:structured_learning}.  First, however we present our approach to feature representation based on convolutional neural networks.

\begin{figure}[ht]
    \centering
    \includegraphics[width=2.5in]{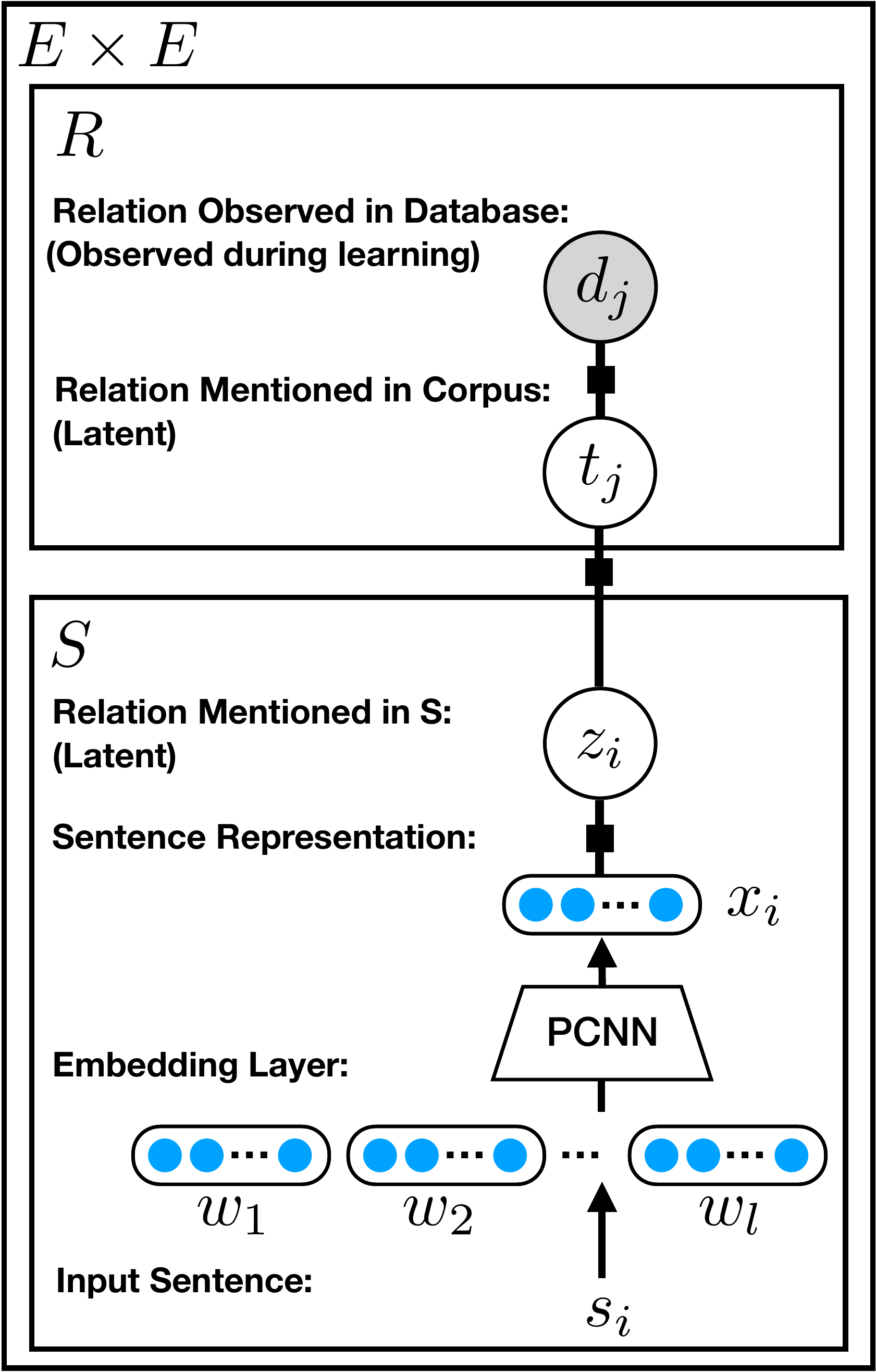}
    \caption{Plate representation of our proposed model.  Plates represent replication; $E \times E$ is the number of entity pairs in the dataset, $S$ is the number of sentences mentioning each entity pair and $R$ is the number of relations.  Arrows represent functions from input to output.  Latent variables are represented as unshaded nodes. Factors over variables are represented as boxes.}
    \label{fig:plate_model}
\end{figure}

\subsection{Mention Representation}
In the following section we review the Piecewise CNN (PCNN) architecture, first proposed by Zeng et. al. \shortcite{zeng-EtAl:2015:EMNLP}, which is used as the basis for our feature representation.


\noindent
{\bf Input Representation:}
A sentence, $s_i$ consisting of $l$ words is represented by two types of embeddings: word embeddings, $E_i$, and position embeddings, $P_i$ relative to the entity pair.  Following Lin et. al. \shortcite{lin-EtAl:2016:P16-1}, word embeddings were initialized by running Word2Vec on the New York Times corpus and later fine-tuned; position embeddings encode the position of the word relative to KB entities, $e_1$ and $e_2$, mentioned in the sentence.
The form of input sentence representation is $w_1, w_2, \cdots, w_l$, where $w_i \in \mathbb{R}^{d}$.  The dimension of embedding at each word position is equal to the word embedding dimension plus two times the position embedding size (one position is encoded for each entity).

\noindent
{\bf Convolution:}
Given an input sentence representation, we perform 1D convolution within a window of length $l$ to extract local features. Assume we have $d_f$ convolutional filters $(F = \{ f_1, f_2, \cdots, f_{d_f} \}, f_i \in \mathbb{R}^{l \times d})$. The output of the $i$-th convolutional filter within the $j$-th window is:

\small
\begin{equation*}
c_{ij} = f_i \cdot w_{j-l+1:j}+b \quad (1 \leq j \leq m+l-1)
\end{equation*}
\normalsize

\noindent
Where $b$ is a bias term. We use zero padding when the window slides out of the sentence boundaries.

\noindent
{\bf Piecewise Max Pooling:}
The output of the convolutional layer $c_i$ is separated into three parts $(c_{i1}, c_{i2}, c_{i3})$ using the positions of the two entities in the sentence.  Max pooling over time is then applied to each of these parts, followed by an elementwise tanh.
The final sentence vector is defined as follows:

\small
\begin{equation*} \label{senPre}
[x]_{ik} = \tanh(\max_j (c_{ikj})) \quad (1 \leq i \leq d_f, 1 \leq k \leq 3)
\end{equation*}
\normalsize

\subsection{Structured Minimally Supervised Learning}
\label{sec:structured_learning}
Our proposed model is based on the PCNN representations described above, in addition to a latent variable model that reasons about missing data and ambiguous relations during learning and is illustrated in Figure \ref{fig:plate_model}.  The embedding for sentence $i$, is used to define a factor over the $i$th input sentence and latent relation mention variable $z_i$:
\[
\phi_\text{PCNN}(s_i, z_i) = e^{x_i \cdot \theta_{z_i}}
\]

\noindent
where $x_i$ is the representation for sentence $s_i$, as encoded by the piecewise CNN.

Another set of factors, $\phi_\text{OR}$, link the sentence-level mention variables, $z_i$, to aggregate-level variables $t_j$, representing whether relation $j$ is mentioned between $e_1$ and $e_2$ in text.  This is modeled using a deterministic OR:
\[
\phi_\text{OR}(\mathbf{z}, t_j) = \mathbf{1}_{\neg t_j \oplus \exists i:j=z_i}
\]

\noindent
where $\mathbf{1}_{x}$ is an indicator function that takes the value 1 when $x$ is true.  The choice of deterministic OR can be interpreted intuitively as follows: if a proposition is true according to $t_j$, then it must be extracted from at least one sentence in the training corpus, on the other hand, if it is false, no sentences in the corpus can mention it.

Finally, we incorporate a set of factors that penalize disagreement between observed relations in the KB, $d_j$, and latent variables $t_j$, which represent whether relation $j$ was extracted from the text.  The penalties for disagreement with the KB are hyperparameters that are adjusted on held-out development data and incorporate entity frequency information from the KB, to model the intuition that more popular entities are less likely to have missing facts:

\small
\[
\phi_\text{A}(t_j,d_j) = 
\begin{cases}
e^{-\alpha_T}, & \text{if } t_j=0 \text{ and } d_j=1 \\
e^{-\alpha_D}, & \text{if } t_j=1 \text{ and } d_j=0 \\
1, & \text{otherwise}
\end{cases}
\]
\normalsize

Putting everything together, the (unnormalized) joint distribution over $\mathbf{t}$, $\mathbf{d}$ and $\mathbf{z}$ conditioned on sentences $\mathbf{s}$ mentioning a dyad is defined as follows:




\small
\begin{equation}
\begin{aligned}
\label{eq:score}
P(\mathbf{d}, \mathbf{t}, \mathbf{z}|\mathbf{s}) \: \propto \: &\prod_{i=1}^{|\mathbf{s}|} \phi_\text{PCNN}(s_i,z_i) \: \times \\ &\Big( \prod_{j=1}^{|\mathbf{r}|} \phi_\text{OR}(\mathbf{z}, t_j) \phi_\text{A}(t_j,d_j)  \Big)^{\mu}  \\
\:=\: &\exp(S_\theta(\mathbf{s}, \mathbf{z},\mathbf{t},\mathbf{d}))
\end{aligned}
\end{equation}
\normalsize

Here, 
$\mu$ is a tunable hyperparameter to adjust impact of the disagreement penalty, and 
$S_\theta(\cdot)$ is the model score for a joint configuration of variables, which corresponds to the log of the unnormalized probability. 

A standard conditional random field (CRF) formulation would optimize model parameters, $\theta$ so as to maximize marginal probability of the observed KB relations, $\mathbf{d}$ conditioned on observed sentences, $\mathbf{s}$:

\small
\[
P(\mathbf{d} | \mathbf{s}) = \sum_{\mathbf{z},\mathbf{t}} P(\mathbf{d}, \mathbf{t}, \mathbf{z}|\mathbf{s})
\]
\normalsize

\noindent
Computing gradients with respect to $P(\mathbf{d}|\mathbf{s})$ (and marginalizing out $\mathbf{z}$ and $\mathbf{t}$) is computationally intractable, so instead we propose an approach that uses maximum-a-posteriori (MAP) parameter learning \citep{taskar2004max} and is inspired by the latent structured SVM \citep{yu2009learning}.


Given a large text corpus in which a set of sentences, $\mathbf{s}$ 
mention a specific pair of entities $(e_1,e_2)$ and a set of relations $\mathbf{d}$ hold between $e_1$ and $e_2$, our goal is to minimize the structured hinge loss: $L_\text{SH}(\theta)=$

\small
\begin{equation}
\max\left\{ 0, 
\genfrac{}{}{0pt}{0}
{
\max\limits_{\mathbf{z}^*_{e},\mathbf{t}^*_{e},\mathbf{d}^*_{e}} \left[S_\theta(\mathbf{s}, \mathbf{z}^*_{e}, \mathbf{t}^*_{e}, \mathbf{d}^*_{e}) + l_\text{Ham}(\mathbf{d}^*_{e}, \mathbf{d}) \right]
}
{
- \max\limits_{\mathbf{z}^*_{g},\mathbf{t}^*_{g}} \left[S_\theta(\mathbf{s}, \mathbf{z}^*_{g}, \mathbf{t}^*_{g}, \mathbf{d}) \right]
}
\right\}
\label{eq:sh}
\end{equation}
\normalsize

\noindent
Where $l_\text{Ham}(\mathbf{d}^*_e, \mathbf{d})$ is the Hamming distance between the bit vector corresponding to the set of observed relations holding between $(e_1,e_2)$ in the KB and those predicted by the model.  Minimizing $L_\text{SH}(\theta)$ can be understood intuitively as adjusting the parameters so that configurations consistent with observed relations in the KB, $\mathbf{d}$, achieve a higher model score than those with a large hamming distance from the observed configuration.  $\mathbf{z}^*_{e}$ corresponds to the most confusing configuration of the sentence-level relation mention variables (i.e. one that has a large score and also a large Hamming loss) 
and $\mathbf{z}^*_{g}$ corresponds to the best configuration that is consistent with the observed relations in the KB.  

This objective can be minimized using stochastic subgradient descent.  Fixing $\mathbf{z}^*_{g}$ and $\mathbf{z}^*_{e}$ to their maximum values in Equation \ref{eq:sh}, subgradients with respect to the parameters can be computed as follows:

{
\small
\medmuskip=0mu
\thinmuskip=0mu
\thickmuskip=0mu
\begin{eqnarray}
\nabla_\theta L_\text{SH}(\theta) & = &
\begin{cases}
\mathbf{0} & \text{if $L_\text{SH}(\theta) \leq 0$}, \\
\nabla_\theta S_\theta(\mathbf{s}, \mathbf{z}^*_{e}, \mathbf{t}^*_{e}, \mathbf{d}^*_{e}) \\ - \nabla_\theta S_\theta(\mathbf{s}, \mathbf{z}^*_{g}, \mathbf{t}^*_{g}, \mathbf{d}) & \text{otherwise}
\end{cases} \\
& = & \begin{cases}
\mathbf{0} & \text{if $L_\text{SH}(\theta) \leq 0$}, \\
\sum_i \nabla_\theta \log \phi_\text{PCNN}(s_{i},z^*_{e,i}) \\
- \sum_i \nabla_\theta \log \phi_\text{PCNN}(s_{i},z^*_{g,i}) & \text{otherwise}
\end{cases}
\label{eq:grad}
\end{eqnarray}
\normalsize
}

Because the second factor of the product in Equation \ref{eq:score} 
does not depend on $\theta$, it is straightforward to compute subgradients of the scoring function, $\nabla S_\theta(\cdot)$, with fixed values of $\mathbf{z}^*_{g}$ and $\mathbf{z}^*_{e}$ using backpropagation (Equation \ref{eq:grad}).

\noindent
{\bf Inference:}
The two inference problems, corresponding to maximizing over hidden variables in Equation \ref{eq:sh} can be solved using a variety of solutions; we experimented with A$^*$ search over left-to-right assignments of the hidden variables.  An admissible heuristic is used to upper-bound the maximum score of each partial hypothesis by maximizing over the unassigned PCNN factors, ignoring inconsistencies.  This approach is guaranteed to find an optimal solution, but can be slow and memory intensive for problems with many variables.  In preliminary experiments on development data, we found that local-search \citep{eisner2006local} using both relation type and mention search operators \citep{liang2010type,ritter13} usually finds an optimal solution and also scales up to large training datasets; we use local search with 30 random restarts to compute argmax assignments for the hidden variables, $\mathbf{z}^*_{g}$ and $\mathbf{z}^*_{e}$, in all our experiments.

\noindent
{\bf Bag-Size Weighting Function:}
Since the search space of the MAP inference problem increases exponentially as the number of hidden variables goes up, it becomes more difficult to find the exact argmax solution using local search, leading to increased noise in the computed gradients. To mitigate the search-error problem in large bags of sentences, we introduce a weighting function based on the bag size as follows:

$$
L_\text{SH}(\theta)^{\prime}=f(s_i) \cdot L_\text{SH}(\theta)
$$

\small
\[
f(s_i) = 
\begin{cases}
1, & \text{if } |\mathbf{s_i}| < \beta_1 \\
\frac{\beta_1}{|\mathbf{s}_i|}, & \text{if } \beta_1 \leq |\mathbf{s}_i| \leq \beta_2 \\
(\frac{\beta_1}{|\mathbf{s}_i|})^2, & \text{otherwise}
\end{cases}
\]
\normalsize

\noindent
where $f(s_i)$ is the bag-size weight for $i$th training entity pair and $\beta_1$/$\beta_2$ are two tunable bag-size thresholds. In Table~\ref{AUCRiedel} and Table~\ref{AUCLin}, we see that this strategy significantly improves performance, especially when training on the larger {\LinDataset} dataset. We also experimented with this method for PCNN+ATT, but found that its performance did not improve.

\section{Experiments}
In Section 2, we presented an approach that combines the benefits of PCNN representations and structured learning with latent variables for minimally supervised relation extraction.  
In this section we present the details of our evaluation methodology and experimental results.

\noindent
{\bf Datasets:}
We evaluate our models on the NYT-Freebase dataset \cite{riedel10modeling} which was created by aligning relational facts from Freebase with the New York Times corpus, and has been used in a broad range of prior work on minimally supervised relation extraction.  Several versions of this dataset have been used in prior work; to facilitate the reproduction of prior results, we experiment with two versions of the dataset used by Riedel et. al. \shortcite{riedel10modeling} (henceforth \RiedelDataset{) }and Lin et. al. \shortcite{lin-EtAl:2016:P16-1} (\LinDataset{}).  Statistics of these datasets are presented in Table \ref{train}.  A more detailed discussion about the differences between datasets used in prior work is also presented in Appendix B.

\begin{table}[t!]
\small
\begin{center}
\begin{tabular}{lcc}
\hline Dataset & \bf \RiedelDataset & \bf \LinDataset \\ 
\bf  & (Riedel et. al. 2010) & (Lin et. al. 2016) \\ 
\hline
Entity pairs & 67,946 & 280,275 \\
Sentences & 126,184 & 523,312 \\
\hline
\end{tabular}
\end{center}
\caption{\label{train}Number of entity pairs and sentences in the training portion of Riedel's {\sc HeldOut} dataset (\RiedelDataset) and Lin's dataset (\LinDataset). }
\end{table}

\noindent
{\bf Hyperparameters:}
Following Lin et. al. \shortcite{lin-EtAl:2016:P16-1}, we utilize word embeddings pre-trained on the NYT corpus using the word2vec tool, other parameters are initialized using the method described by Glorot and Bengio \shortcite{glorot2010understanding}. The Hoffmann et. al. sentential evaluation dataset is split into a development and test set and grid search on the development set was used to determine optimal values for the learning rate $\lambda$ among $\{0.001, 0.01\}$, 
KB disagreement penalty scalar $\mu$ among $\{100, 200, \cdots,  2000\}$ and $\beta_1$/$\beta_2$ bag size threshold for the weighting function among $\{10, 15, \cdots,  40\}$.  Other hyperparameters with fixed values are presented in Table \ref{paraSetting}.

\begin{table}[t!]
\small
\begin{center}
\begin{tabular}{|c|c|}
\hline

Window length $l$ & 3 \\
Number of convolutional filters $d_f$ & 230 \\
Word embedding dimension $d_w$ & 50 \\
Position embedding dimension $d_p$ & 5 \\
Batch size $B$ & 1 \\

\hline
\end{tabular}
\end{center}
\caption{\label{paraSetting}Untuned hyperparameters in our experiments.}
\end{table}



\noindent
\textbf{Neural Baselines:}
To demonstrate the effectiveness of the our approach, we compare against col-less universal schema \cite{verga-EtAl:2016:N16-1} in addition to the PCNN+ATT model of Lin et. al. \shortcite{lin-EtAl:2016:P16-1}.
After training the Lin et. al. model to predict observed facts in the KB, we use its attention layer to make mention-level predictions as follows:

\small
\begin{equation*}
p(r_j|x_i) = \frac{\text{exp}(r_j \cdot x_i)}{\sum_{k=1}^{n_r} \text{exp}(r_k \cdot x_i)}
\end{equation*}
\normalsize
\noindent
Where $r_j$ indicates the vector representation of the $j$th relation.

\noindent
\textbf{Structured Baselines:}
In addition to initializing convolutional filters used in the $\phi_{\text{PCNN}}(\cdot)$ factors randomly and performing structured learning of representations as in Equation \ref{eq:grad}, we also experimented with variants of MultiR and DNMAR, which are based on the structured perceptron \citep{collins2002discriminative}, using fixed sentence representations: both traditional sparse feature representations, in addition to pre-trained continuous representations generated using our best-performing reimplementation of PCNN+ATT.  For the structured perceptron baselines, we also experimented with variants based on MIRA \citep{crammer2003ultraconservative}, which we found to provide consistent improvements.  More details are provided in Appendix A.


\subsection{Sentential Evaluation}
\label{sec:results}


In this work, we are primarily interested in mention-level relation extraction.  
For our first set of experiments (Tables \ref{AUCRiedel} and \ref{AUCLin}), we use the manually annotated dataset created by \cite{hoffmann-EtAl:2011:ACL-HLT2011}. 
Note that sentences in the Hoffman et. al. dataset were selected from the output of systems used in their evaluation, so it is possible there are high confidence predictions made by our systems that are not present. 
Therefore, we further validate our findings, by performing a manual inspection of the highest confidence predictions in Table \ref{P@N}.

\noindent
{\bf \RiedelDataset{} Results:}
As illustrated in Table~\ref{AUCRiedel}, simply applying structured models (MultiR and DNMAR) with pre-trained sentence representations performs competitively.  
MIRA provides consistent improvements for both sparse and dense representations. 
PCNN+ATT outperforms most latent-variable models on the sentential evaluation, we found this result to be surprising as the model was designed for extracting proposition-level facts. 
Col-less universal schema does not perform very well in this evaluation; this is likely due to the fact that it was developed for the KBP slot filling evaluation \cite{ji2010overview}, and only uses the part of a sentence between two entities as an input representation, which can remove important context. Our proposed model, which jointly learns sentence representations using a structured latent-variable model that allows for the possiblity of missing data, achieves the best overall performance; its improvements over all baselines were found to be statistically significant according to a paired bootstrap test \cite{efron1994introduction,berg2012empirical}.\footnote{p-value is less than 0.05.}

\begin{table*}[t!]
\small
\begin{center}
\begin{tabular}{llccc}
\hline Model & Name & DEV & TEST\\ \hline
\multirow{8}{2cm}{Fixed Sentence Representations} & MultiR\_sparse \cite{hoffmann-EtAl:2011:ACL-HLT2011} & 66.2 & 63.2 \\
& MultiR\_sparse\_MIRA & 75.3 & 71.6 \\
& MultiR\_continuous & 74.2 & 68.7 \\
& MultiR\_continuous\_MIRA & 80.3 & 72.5 \\
& DNMAR\_sparse \cite{ritter13} & 77.9 & 70.1 \\
& DNMAR\_sparse\_MIRA & 77.5 & 72.1 \\
& DNMAR\_continuous & 80.2 & 70.0 \\
& DNMAR\_continuous\_MIRA & 82.2 & 74.2 \\ \hline
\multirow{2}{2.5cm}{Jointly Learned Representations}& \OurSystem{} & 82.4 & 83.9 \\ 
& \OurSystem{} (bag-size weighting function) & \bf 85.4 & \bf 86.0 \\  \hline
\multirow{3}{2cm}{Baselines} & col-less universal schema \cite{verga-EtAl:2016:N16-1} & 63.4 & 61.1\\
& PCNN+ATT (Lin et al. \shortcite{lin-EtAl:2016:P16-1} code) & 81.4 & 76.4\\
& PCNN+ATT (our reimplementation with parameter tuning) & 83.6 & 78.4\\

\hline
\end{tabular}
\end{center}
\caption{\label{AUCRiedel}AUC of sentential evaluation precision / recall curves for all models trained on \RiedelDataset{}. Continuous sentence representation works as well as human-engineered sentence representation, and MIRA consistently helps structured perceptron training. PCNN+ATT performs competitively while our \OurSystem{} (weighted) is statistically significantly better (p-value of bootstrap is less than 0.05)}
\end{table*}

\noindent
{\bf \LinDataset{} Results:}
When training on the larger dataset provided by Lin et. al. \shortcite{lin-EtAl:2016:P16-1}, linguistic features are not available, so only neural representations are included in our evaluation. 
As illustrated in Table~\ref{AUCLin}, \OurSystem{} also achieves the best performance when training on the larger dataset; its improvements over the baselines are statistically significant.  
The AUC of most models decreases on the Hoffmann et. al. sentential dataset when training on \LinDataset{}.
This is not surprising, because the Hoffmann et. al. dataset is built by sampling sentences from positive predictions of models trained on \RiedelDataset{}; changing the training data causes a difference in the ranking of high-confidence predictions for each model, leading to the observed decline in performance against the Hoffmann et. al. dataset.  To further validate our findings, we also manually inspect the models' top predictions as described below.

\begin{table*}[t!]
\small
\begin{center}
\begin{tabular}{llccc}
\hline Model & Name & DEV & TEST\\ \hline
\multirow{4}{2cm}{Fixed Sentence Representations} & MultiR\_continuous & 72.4 & 66.7 \\
& MultiR\_continuous\_MIRA & 74.6 & 73.4 \\
& DNMAR\_continuous & 73.1 & 68.0 \\
& DNMAR\_continuous\_MIRA & 75.6 & 68.7 \\ \hline
\multirow{2}{2.5cm}{Jointly Learned Representations}& \OurSystem{} & 78.1 & 75.4 \\ 
& \OurSystem{} (bag-size weighting function) & \bf 82.9 & \bf 83.1 \\ \hline
\multirow{3}{2cm}{Baselines} & col-less universal schema \cite{verga-EtAl:2016:N16-1} & 60.3 & 57.5  \\
& PCNN+ATT (Lin et al. \shortcite{lin-EtAl:2016:P16-1} code) & 67.9 & 72.1 \\
& PCNN+ATT (our reimplementation with parameter tuning) & 78.2 & 74.8\\

\hline
\end{tabular}
\end{center}
\caption{\label{AUCLin}AUC of sentential evaluation precision / recall curves for all models trained on \LinDataset{}. Our proposed \OurSystem{} (weighted) still performs the best, and the advantage over baselines is also statistically significant (p-value of bootstrap is less than 0.05).}
\end{table*}

\noindent
{\bf Manual Evaluation:}
Because the Hoffmann et. al. sentential dataset does not contain the highest confidence predictions, we also manually inspected each model's top 500 predictions for the most frequent 4 relations, and report precision @ N to further validate our results.  As shown in Table~\ref{P@N}, for \RiedelDataset{}, PCNN+ATT performs comparably on {\tt /location/contains}\footnote{{\tt /location/contains} is the most frequent relation in the Hoffmann et. al. dataset.} and {\tt /person/company}, whereas our model has a considerable advantage on the other two relations.  For \LinDataset{}, our model performs consistently better on all four relations compared with PCNN+ATT.  When training on the larger \LinDataset{} dataset, we observe trend of increasing mention-level P@N for \OurSystem{}, however the performance of PCNN+ATT appears to decrease.
We investigate this phenomenon further below.

\begin{table}[t!]
\small
\begin{center}
\setlength\tabcolsep{2pt}
\scalebox{1}{
\begin{tabular}{lccc}
\hline 

\multirow{2}{*}{Relation} & \multirow{2}{*}{N} & \multirow{2}{*}{PCNN+ATT} & \OurSystem{} \\ 
&  & & (weighted) \\ \hline

\multicolumn{4}{c}{\RiedelDataset{}} \\ \hline
\multirow{2}{*}{/location/contains} & 100 & \bf1.00 & 0.99\\
& 500 & 0.97 & \bf 0.98 \\\hline
\multirow{2}{*}{/person/place\_lived} & 100 & 0.76 & \bf0.98\\ 
& 500 & 0.63 & \bf 0.78 \\\hline
\multirow{2}{*}{/person/nationality} & 100 & 0.62 & \bf0.89\\ 
& 500 & 0.43 & \bf 0.54 \\\hline
\multirow{2}{*}{/person/company} & 100 & 0.98 & 0.98\\ 
& 500 & 0.72 & \bf 0.78 \\ \hline
\multicolumn{4}{c}{\LinDataset{}} \\ \hline
\multirow{2}{*}{/location/contains} & 100 & 0.98 & \bf0.99\\
& 500 & 0.82 & \bf0.99 \\\hline
\multirow{2}{*}{/person/place\_lived} & 100 & 0.58 & \bf0.98\\ 
& 500 & 0.57 & \bf0.84 \\\hline
\multirow{2}{*}{/person/nationality} & 100 & 0.70 & \bf0.91\\ 
& 500 & 0.35 & \bf0.56 \\\hline
\multirow{2}{*}{/person/company} & 100 & 0.59 & \bf0.95\\ 
& 500 & 0.40 & \bf0.68 \\ \hline 
\end{tabular}
}
\end{center}

\caption{\label{P@N}Top: P@N of 4 most frequent relations for models trained on \RiedelDataset{}. Bottom: P@N of 4 most frequent relations  for models trained on \LinDataset{}. Both models can perform well on {\tt /location/contains} relation while \OurSystem{} (weighted) is consistently better over other relations.}

\end{table}

\begin{table}[t!]
\small
\begin{center}
\setlength\tabcolsep{2pt}
\begin{tabular}{lccc}
\hline Category & True & False  & Total \\ \hline
\multicolumn{4}{c}{DEV} \\ \hline
In-Freebase & 102 & 180 & 282 \\
Out-Of-Freebase & 58 & 96 & 154 \\ \hline
\multicolumn{4}{c}{TEST} \\ \hline
In-Freebase & 113 & 192 & 305 \\
Out-Of-Freebase & 41 & 99 & 140 \\

\hline
\end{tabular}
\end{center}
\caption{\label{SenDistribution}Top: Sentence distribution in Hoffmann et. al. \shortcite{hoffmann-EtAl:2011:ACL-HLT2011} sentential evaluation DEV dataset. Bottom: Sentence distribution in Hoffmann et. al. \shortcite{hoffmann-EtAl:2011:ACL-HLT2011} sentential evaluation TEST dataset. There are substantial Out-Of-Freebase mentions which are manually labelled as correct relational mentions.}
\end{table}


\begin{table}[t!]
\small
\begin{center}
\setlength\tabcolsep{2pt}
\begin{tabular}{llccc}
\hline Model & Dataset & InFB & OutFB \\ \hline
\multicolumn{4}{c}{DEV} \\ \hline
\multirow{3}{*}{PCNN+ATT} & \RiedelDataset & 78.2 & 89.6 \\
 & \LinDataset & 77.1 & 77.0 \\ 
 & Change &  -1.1 & \bf -12.6 \\ \hline
 \multirow{3}{*}{\OurSystem (weighted)} & \RiedelDataset & 81.3 & 90.4 \\
 & \LinDataset & 77.7 & 90.6 \\ 
 & Change & \bf -3.6 & +0.2 \\ \hline

\multicolumn{4}{c}{TEST} \\ \hline
\multirow{3}{*}{PCNN+ATT} & \RiedelDataset & 78.7 & 75.9 \\
 & \LinDataset & 81.9 & 56.8 \\ 
 & Change & +3.2 & \bf -19.1 \\ \hline
 \multirow{3}{*}{\OurSystem (weighted)} & \RiedelDataset & 85.9 & 85.4 \\
 & \LinDataset & 83.1 & 81.5 \\ 
 & Change & -2.8 & \bf -3.9 \\ \hline

\end{tabular}
\end{center}
\caption{\label{InOutFBAUC} Top: Comparison of AUCs of In-Freebase and Out-Of-Freebase mentions on sentential DEV set for PCNN+ATT and \OurSystem{} (weighted) with two datasets. Bottom: Comparison of AUCs of In-Freebase and Out-Of-Freebase mentions on sentential TEST set for PCNN+ATT and \OurSystem{} (weighted) with two datasets. PCNN+ATT has significant drops about Out-Of-Freebase mentions on both sentential DEV and TEST set after training on the larger {\LinDataset}  which explains why its overall AUC performances goes down while \OurSystem{} (weighted) does not have such problem.}
\end{table}

\noindent
{\bf Performance at Extracting New Facts:}
\label{sec:evaluation_analysis}
To explain PCNN+ATT's drop in mention-level performance after training on the larger \LinDataset{} dataset, our hypothesis is that the larger KB-supervised dataset not only contains more true positive training examples but also more false negative examples.  This biases models toward predicting facts about popular entities, which are likely to exist in Freebase.
To provide evidence in support of this hypothesis, we divide the manually annotated dataset from Hoffmann et. al. into two categories: mentions of facts found in Freebase, and those that are not; this distribution is presented in the Table~\ref{SenDistribution}. 
In Table~\ref{InOutFBAUC}, we present a breakdown of model performance on these two subsets.  For PCNN+ATT, although the AUC of in-Freebase mentions on the test set increases after training on the larger \LinDataset{}, its Out-Of-Freebase AUC on both dev and test sets drops significantly, which clearly illustrates the problem of increasing false negatives during training.  In contrast, our model, which explicitly allows for the possibility of missing data in the KB during learning, has relatively stable performance on the two types of mentions, as the amount of weakly-supervised training data is increased.



\subsection{Held-Out Evaluation}
\label{sec:held_out_evaluation}
In Section \ref{sec:results}, we evaluated the results of minimally supervised approaches to relation extraction by comparing extracted mentions against human judgments.
An alternative approach, which has been used in prior work, is to evaluate a model's performance by comparing predictions against held out facts from a KB.  
Taken in isolation, this approach to evaluation can be misleading, because it penalizes models that extract many new facts that do not already appear in the knowledge base.  This is  undesirable, because the whole point of an information extraction system is to extract {\em new} facts that are not already contained in a KB.
Furthermore, sentential extraction has the benefit of providing clear provenance for extracted facts, which is crucial in many applications.  Having mentioned these limitations of the held-out evaluation metrics, however, we now present results using this approach to facilitate comparison to prior work.

\begin{figure}[h]
\includegraphics[width=7cm]{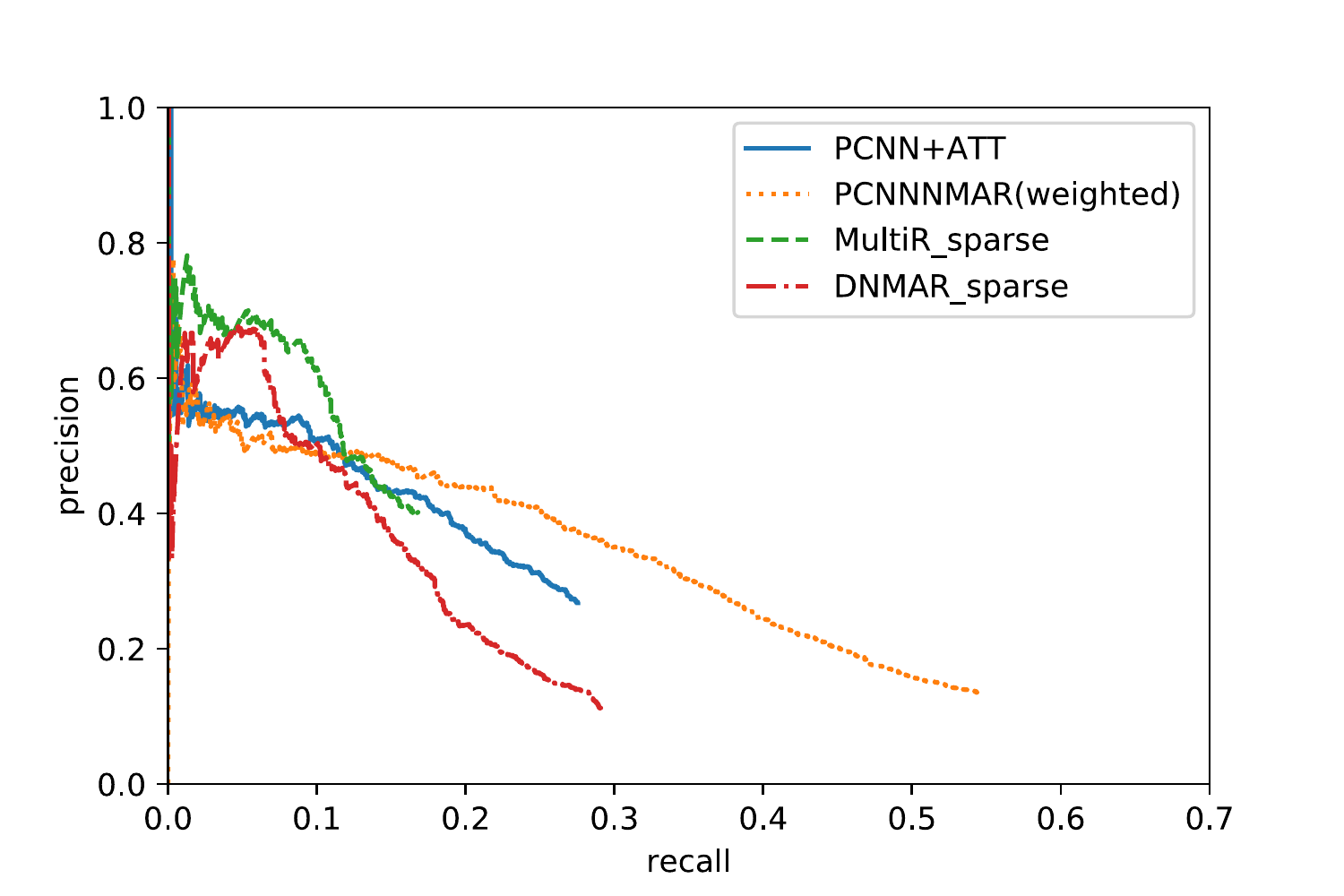}
\caption{\label{Riedel}Held-out evaluation precision / recall curves for PCNN+ATT, MultiR, DNMAR and our proposed model \OurSystem{} (weighted) on \RiedelDataset{}.}

\end{figure}
\begin{figure}[h]
\includegraphics[width=7cm]{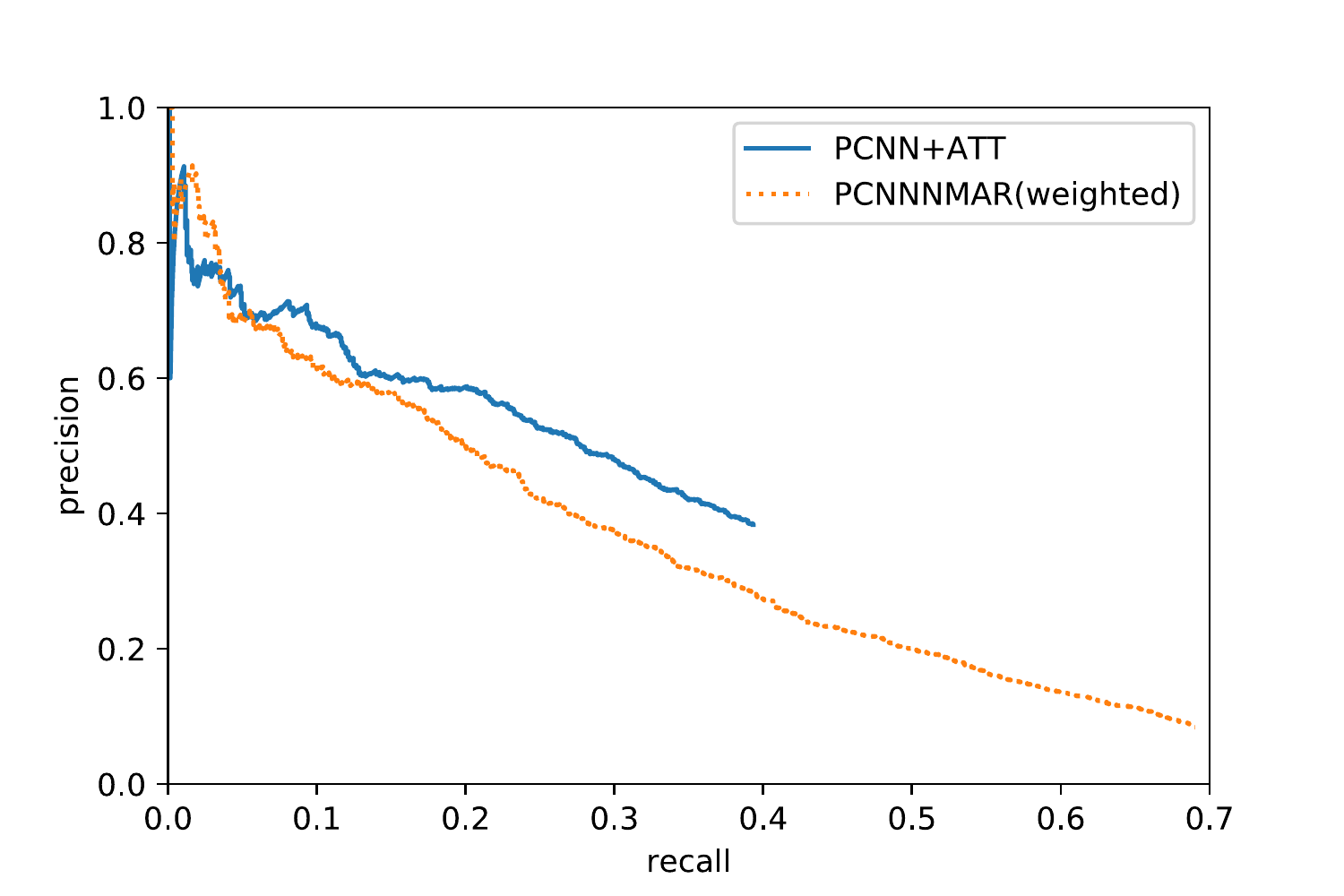}
\caption{\label{Lin}Held-out evaluation precision / recall curves for all NN-based models on \LinDataset{}.}
\end{figure}

Figure \ref{Riedel} presents precision-recall curves against held out facts from Freebase comparing \OurSystem{} to several baselines and Figure \ref{Lin} presents results on the larger \LinDataset{} dataset.  All models perform better according to the held out evaluation metric when training on the larger dataset, which is consistent with our hypothesis, presented at the end of Section \ref{sec:evaluation_analysis}.  Our structured model with learned representations, \OurSystem{} (weighted), has lower precision when recall is high. This also fits with our hypothesis, as systems that explicitly model missing data will extract many correct facts that do not appear in the KB, resulting in an under-estimate of precision according to this metric.



\section{Related Work}

\noindent
{\bf Knowledge Base Population:} There is a long line of prior work on learning to extract relational information from text using minimal supervision. Early work on semantic bootstrapping \cite{Hearst:1992:AAH:992133.992154,brin1998extracting,agichtein2000snowball,carlson2010toward,gupta2014improved,qu2018weakly}, applied an iterative procedure to extract lexical patterns and relation instances.  These systems tend to suffer from the problem of semantic drift, which motivated work on distant supervision \cite{craven1999constructing,snyder2007database,wu2007autonomously,mintz2009distant}, that explicitly minimizes standard loss functions, against observed facts in a knowledge base.  
The TAC KBP Knowledge Base Population task was a prominent shared evaluation of relation extraction systems \citep{ji2010overview,surdeanu2013overview,surdeanu2010simple,surdeanu2012multi}.  
Recent work has explored a variety of new neural network architectures for relation extraction \citep{wang-EtAl:2016:P16-12,zhang2017position,yu2015combining}, experimenting with alternative sentence representations in our framework is an interesting direction for future work.
Recent work has also shown improved performance by incorporating supervised training data on the sentence level \citep{angeli2014combining,beltagy2018improving}, in contrast our approach does not make use of any sentence-level labels during learning and therefore relies on less human supervision.
Finally, prior work has explored a variety of methods to address the issue of noise introduced during distant supervision \cite{wu2017adversarial,yaghoobzadeh2017noise,P18-1046}.



Another line of work has explored open-domain and unsupervised methods for IE \citep{yao2011structured,ritter12,stanovsky2015open,huang2016liberal,weber2017event}.
Universal schemas \citep{riedel2013relation} combine aspects of minimally supervised and unsupervised approaches to knowledge-base completion by applying matrix factorization techniques to multi-relational data \citep{nickel2011three,bordes2013translating,chang2014tensor}.  Rows of the matrix typically model pairs of entities, and columns represent relations or syntactic patterns (i.e., syntactic dependency paths observed between the entities).


\noindent
{\bf Structured Learning with Neural Representations:}
Prior work has investigated the combination of structured learning with learned representations for a number of NLP tasks, including parsing \cite{weiss2015structured,durrett-klein:2015:ACL-IJCNLP,andor2016globally}, named entity recognition \cite{cherry2015unreasonable,ma2016end,lample2016neural} and stance detection \citep{li2018structured}.  We are not aware of any previous work that has explored this direction on the task of minimally supervised relation extraction; we believe structured learning is particularly crucial when learning from minimal supervision to help address the issues of missing data and overlapping relations.



\section{Conclusions}
In this paper we presented a hybrid approach to minimally supervised relation extraction that combines the benefits of structured learning and learned representations.  Extensive experiments show that by performing inference during the learning procedure to address the issue of noise in distant supervision, our proposed model achieves state-of-the-art performance on minimally supervised mention-level relation extraction.



\section*{Acknowledgments}
Funding was provided by the National Science Foundation under Grant No. IIS-1464128, the Defense Advanced Research Projects Agency (DARPA) via the U.S. Army Research Office (ARO) and under Contract Number W911NF-17-C-0095 and the Office of the Director of National Intelligence (ODNI) and Intelligence Advanced Research Projects Activity (IARPA) via the Air Force Research Laboratory (AFRL) contract number FA8750-16-C0114, in addition to an Amazon Research Award and an NVIDIA GPU grant.  The content of the information in this document does not necessarily reflect the position or the policy of the Government, and no official endorsement should be inferred. The U.S. Government is authorized to reproduce and distribute reprints for government purposes notwithstanding any copyright notation here on.

\bibliography{naaclhlt2019}
\bibliographystyle{acl_natbib}

\appendix

\section{MIRA}
Prior work on minimally supervised structured learning has made use of sparse feature representations in combination with perceptron-style parameter updates.  We found these updates result in poor performance on held-out development data, however, when using fixed, pre-trained continuous sentence representations.  Perhaps this is not surprising because, intuitively, the margin of the dataset is likely to be smaller when using lower dimensional, continuous representations, leading to a larger mistake bound for convergence of the perceptron.  To address this, we applied the the {\bf M}argin {\bf I}nfused {\bf R}elaxation {\bf A}lgorithm  \cite{crammer2003ultraconservative}, as described below.  In Section 3.1, we show empirically that MIRA is crucial for achieving good performance when using continuous representations, and consistently improves performance when using sparse features as well.

As discussed above, we have $\hat{\mathbf{z}}^\text{KB}$ the most likely sentence extractions conditioned on the KB and $\hat{\mathbf{z}}$, the MAP assignment to $\mathbf{z}$, ignoring the KB. MIRA updates parameters of the PCNN factors as follows:



\small
\begin{eqnarray*}
\theta_j & = & \theta_j + \tau \cdot \left( F_j(x_i, \hat{z}_i^\text{KB}) - F_j(x_i, \hat{z}_i) \right) \\
\end{eqnarray*}
\normalsize

\noindent
here $\tau$ is an adaptive learning rate that scales the update to the smallest step size that achieves 0 loss on each mention-level classification:

\small
\begin{eqnarray*}
\tau & = & \min \bigg( C, \frac{1-\theta \cdot \left( F(x_i, \hat{z}_i^\text{KB}) -  F(x_i, \hat{z}_i) \right)}{{2 \lvert \lvert x_i \rvert \rvert}^2} \bigg)
\end{eqnarray*}
\normalsize

\noindent
$\theta$ is the concatenation of parameters $\theta_j$ across relations $j$, and similarly $F(\cdot)$ is the concatenation of PCNN features across relations.  $C$ is a hyper-parameter that truncates large steps and helps to prevent overfitting.

\section{Differing Versions of the NYT-Freebase Corpus Used in Prior Work}
We evaluate our models on the NYT-Freebase dataset \cite{riedel10modeling} which was created by aligning relational facts from Freebase with the New York Times corpus, and has been used in a broad range of prior work on minimally supervised relation extraction. Originally, Riedel et. al. created two separate datasets for their {\sc HeldOut} and {\sc Manual} evaluations. In the {\sc HeldOut} dataset, Freebase entity pairs are divided into two parts, one for training and one for testing. Training dyads are aligned to the 2005-2006 portion of the NYT corpus while testing dyads are aligned to the year 2007. In the {\sc Manual} evaluation data, {\bf all} Freebase entity pairs are matched against the 2005-2006 articles and used as training instances. Testing data in the Riedel et. al. {\sc Manual} evaluation consists of dyads found within sentences in the 2007 NYT articles, for which at least one entity does not appear in Freebase; their models' predictions on this data were annotated manually. 
The Riedel et. al. data splits ensure it is not possible to have overlapping train/test entity pairs in either the {\sc HeldOut} or {\sc Manual} evaluation.

\begin{figure}[h]
\includegraphics[width=8cm]{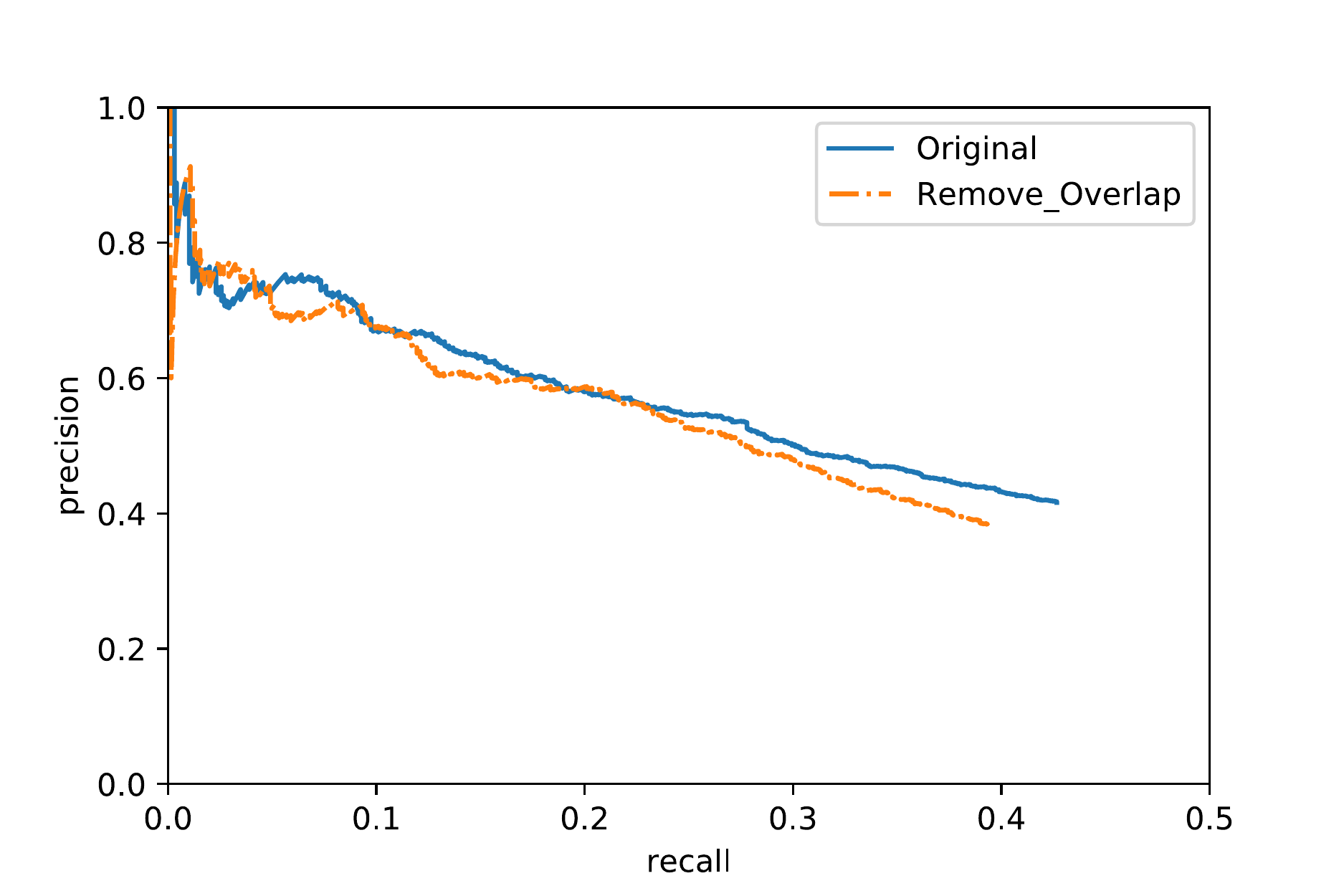}
\caption{\label{RemoveOverlapAgg}Held-out evaluation precision / recall curves for PCNN+ATT model on original \LinDataset{} and its shared-entity-pairs-removed version.}
\end{figure}

As neural models with many parameters typically benefit significantly from larger quantities of training data, Lin et. al. \shortcite{lin-EtAl:2016:P16-1} added training data from the Riedel et. al. {\sc Manual-Train} dataset into their training dataset. This modification of the training data leads to overlap in the entity pairs in the Lin et. al. training/test split.  We found 11,424 entity pairs appearing in both training and test sets, however no sentences appear in both the training and test sets, as the matched NYT articles came from different time periods.\footnote{We downloaded the Lin et. al. \shortcite{lin-EtAl:2016:P16-1} dataset from the associated Github repository (\url{https://github.com/thunlp/NRE}) on June, 2017.  The repository was updated in March and May 2018, addressing the overlapping-entity-pairs issue using the same approach described in our paper.}  In all our evaluations we remove these overlapping entity pairs from the training set, to ensure the models are not simply memorizing KB facts that appear in the training data.
Figure~\ref{RemoveOverlapAgg} shows that after removing these shared entity pairs from the training data, performance of the Lin et. al. PCNN+ATT model does not change very much when evaluating against held out facts from Freebase.

\begin{figure}[h]
\includegraphics[width=8cm]{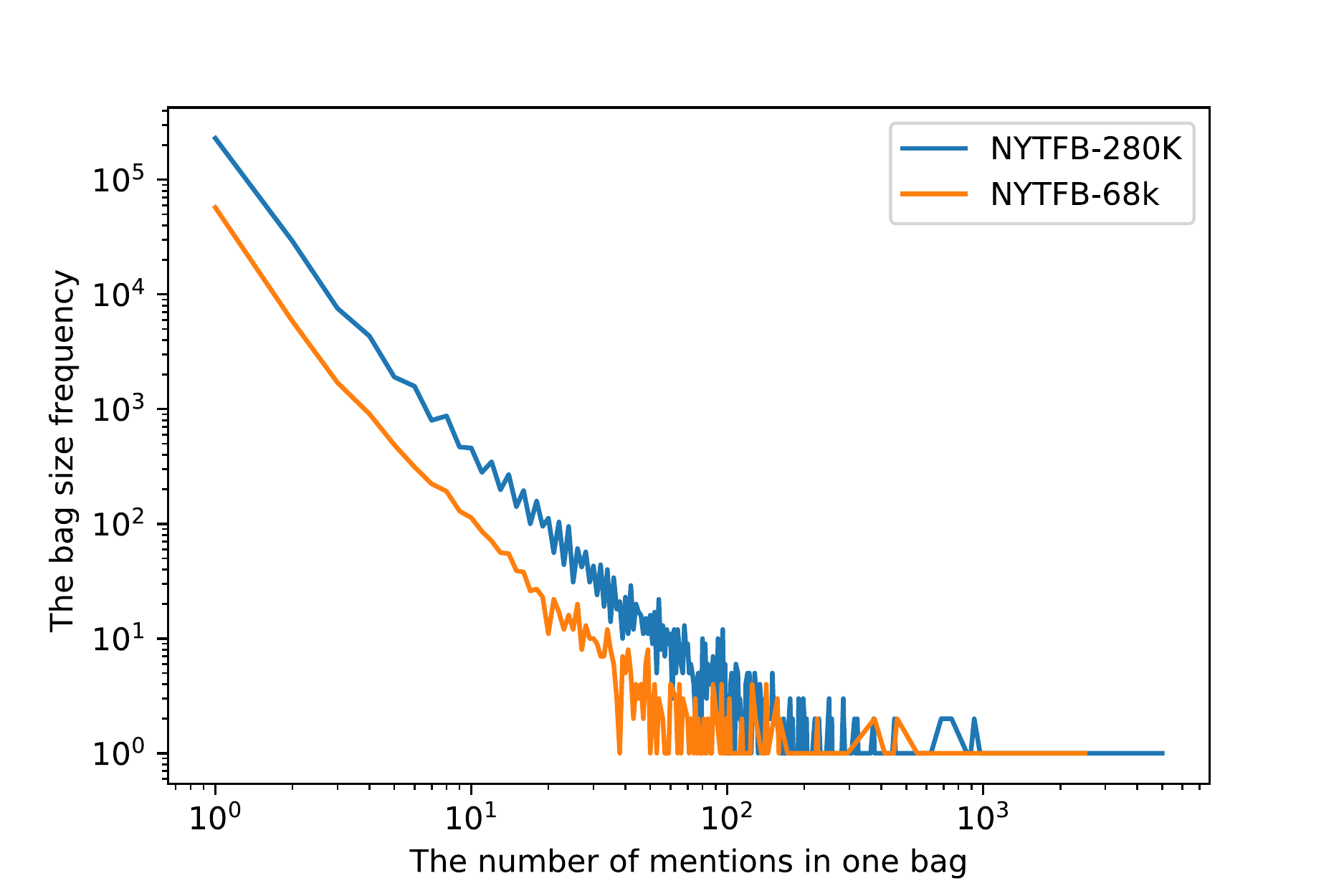}
\caption{\label{bagSizeCompare}Distribution of bag size in the training set of the \RiedelDataset{} and \LinDataset{}.}
\end{figure}

We name two versions of the NYT-Freebase dataset according to the number of training entity pairs they include. Table~\ref{train} shows that \LinDataset{} training set has around 4 times the number of sentences and entity pairs as \RiedelDataset{}, and the proportions of multi-sentence entity pairs in \LinDataset{} is higher. In Table~\ref{riedelLinRel}, we can see that the distribution of relations in the two datasets are comparable, but \LinDataset{} has much more entity pairs for each relation. Also, Figure~\ref{bagSizeCompare} tells us that \LinDataset{} has a wider bag-size range and more large training bags. 


 
\begin{table}[t!]
\small
\begin{center}
\begin{tabular}{lcc}
\hline Dataset & \bf \RiedelDataset{} & \bf \LinDataset{} \\ 
\bf  & (Riedel et. al. 2010) & (Lin et. al. 2016) \\ 
\hline

Entity pairs & 67,946 & 280,275 \\
Sentences & 126,184 & 523,312 \\
Distinct sent. & 96,340 & 340,970 \\
Relations & 52 & 53 \\
\hline
\end{tabular}
\end{center}
\caption{\label{train}Number of entity pairs and sentences in the training portion of Riedel's {\sc HeldOut} dataset (\RiedelDataset) and Lin's dataset (\LinDataset). }
\end{table}

\begin{table}[!h]
\centering
\small
\resizebox{.48\textwidth}{!}{
\begin{tabular}{lcccc}
\hline
\multirow{2}{*}{Relation} & \multicolumn{2}{c}{\bf \RiedelDataset} & \multicolumn{2}{c}{\bf \LinDataset} \\
& \# EPs & percent & \# EPs & percent \\ \hline

NA & 63596 & 93.12 & 263372 & 93.52\\
/location/contains & 2147 & 3.14 & 7760 & 2.76\\
/person/place\_lived & 581 & 0.85 & 2300 & 0.86\\ 
/person/nationality & 436 & 0.64 & 2553 & 0.87\\ 
/person/place\_of\_birth & 370 & 0.54 & 1400 & 0.49\\
/person/company & 357 & 0.52 & 1417 & 0.50\\ \hline

\end{tabular}
}
\caption{\label{riedelLinRel}Distribution of the most frequent relations in the training set of \RiedelDataset{} and \LinDataset{}.}
\end{table}

\section{Variations on Structured Hinge Loss}
\label{sec:loss_function_comparison}

\begin{table}[t!]
\small
\begin{center}
\setlength\tabcolsep{2pt}
\begin{tabular}{lccc}
\hline Method &  &  DEV &  TEST \\ \hline

\multirow{2}{*}{0/1 loss} & normal & 82.6 & 82.8 \\ 
& weighted & 83.9 & 81.3 \\ \hline
\multirow{2}{*}{relation-level} & normal & 83.9 & 83.1  \\ 
 & weighted & 84.6 & 81.1  \\ \hline
\multirow{2}{*}{mention-level} & normal & 82.4 & 83.9  \\ 
 & weighted & \bf 85.4 & \bf 86.0  \\

\hline
\end{tabular}
\end{center}
\caption{\label{threeLoss}AUC of sentential evaluation precision / recall curves for \OurSystem{} with three loss functions trained on \RiedelDataset{}. Mention-level hamming loss has some advantages over other two loss functions.}
\end{table}

Since we use the hinge loss as the loss function in our proposed \OurSystem{} model, the way that the hamming loss is calculated  decides how we solve the argmax problem in loss-augmented search. In our experiments, we explore three ways to compute the loss: 0/1 loss, relation-level hamming loss and mention-level hamming loss. Table~\ref{threeLoss} shows that mention-level hamming loss has obvious advantage on AUC performance over other two methods. Although theoretically relation-level hamming loss should be better, it is really hard to find the exact argmax solution in loss-augmented inference with local search while we can easily get it with mention-level hamming loss.

\end{document}